\documentclass[letterpaper,10pt,conference]{ieeeconf}
\IEEEoverridecommandlockouts
\usepackage{adathinkdrive_ieee}
\hypersetup{pdftitle={AdaThinkDrive: Adaptive Thinking via Reinforcement Learning for Autonomous Driving},pdfauthor={Yuechen Luo et al.}}
\title{\LARGE\bfseries AdaThinkDrive: Adaptive Thinking via Reinforcement Learning\\for Autonomous Driving}
\author{\small
Yuechen Luo\textsuperscript{1,3*}, Fang Li\textsuperscript{2,3*}, Shaoqing Xu\textsuperscript{2,3*,\textdaggerdbl}, Zhiyi Lai\textsuperscript{3}, Lei Yang\textsuperscript{4}, 
Qimao Chen\textsuperscript{1,3}, Ziang Luo\textsuperscript{1}, Zixun Xie\textsuperscript{3,5},\\[3pt] Shengyin Jiang\textsuperscript{3}, Jiaxin Liu\textsuperscript{1,3}, 
Long Chen\textsuperscript{3}, Bing Wang\textsuperscript{3}, Ken Deng\textsuperscript{6}, Zhi-xin Yang\textsuperscript{2,\ding{41}}
\\[5pt]
\textbf{Code:} \url{https://github.com/luo-yc17/AdaThinkDrive}
\thanks{\textsuperscript{1}Tsinghua University, \textsuperscript{2}University of Macau, \textsuperscript{3}Xiaomi EV.}
\thanks{\textsuperscript{4}Nanyang Technological University, \textsuperscript{5}Peking University.}
\thanks{\textsuperscript{6}Beijing University of Posts and Telecommunications.}
\thanks{Email: \href{mailto:luo-yc24@mails.tsinghua.edu.cn}{luo-yc24@mails.tsinghua.edu.cn}.}
\thanks{*Equal contribution.\quad \textdaggerdbl\ Project Leader.\quad \ding{41} Corresponding author.}
}
\begin{document}
\maketitle
\thispagestyle{empty}
\pagestyle{plain}
\begin{abstract}
While reasoning technology like Chain-of-Thought (CoT) has been widely adopted in Vision-Language-Action (VLA) models, it demonstrates promising capabilities in end-to-end autonomous driving. However, recent efforts to integrate CoT reasoning often fall short in simple scenarios, introducing unnecessary computational overhead without improving decision quality. To address this, we propose \textbf{AdaThinkDrive}, a novel VLA framework with a dual-mode reasoning mechanism inspired by fast and slow thinking. First, our framework is pretrained on large-scale autonomous driving (AD) scenarios using both question-answering (QA) and trajectory datasets to acquire world knowledge and driving commonsense. During supervised fine-tuning (SFT), we introduce a two-mode dataset—fast answering (w/o CoT) and slow thinking (with CoT), enabling the model to distinguish between scenarios that require reasoning. Furthermore, an \textit{Adaptive Think Reward} strategy is proposed in conjunction with the \textit{Group Relative Policy Optimization (GRPO)}, which rewards the model for selectively applying CoT by comparing trajectory quality across different reasoning modes. Extensive experiments on the Navsim benchmark show that AdaThinkDrive achieves a PDMS of 90.3, surpassing the best vision-only baseline by 1.7 points. Moreover, ablations show that AdaThinkDrive surpasses both the never-Think and always-Think baselines, improving PDMS by 2.0 and 1.4, respectively. It also reduces inference time by 14\% compared to the always-Think baseline, demonstrating its ability to balance accuracy and efficiency through adaptive reasoning.
\end{abstract}
\section{Introduction}

\begin{figure}[t!]
    \centering
    \begin{subfigure}{\linewidth}
        \centering
        \includegraphics[width=\linewidth]{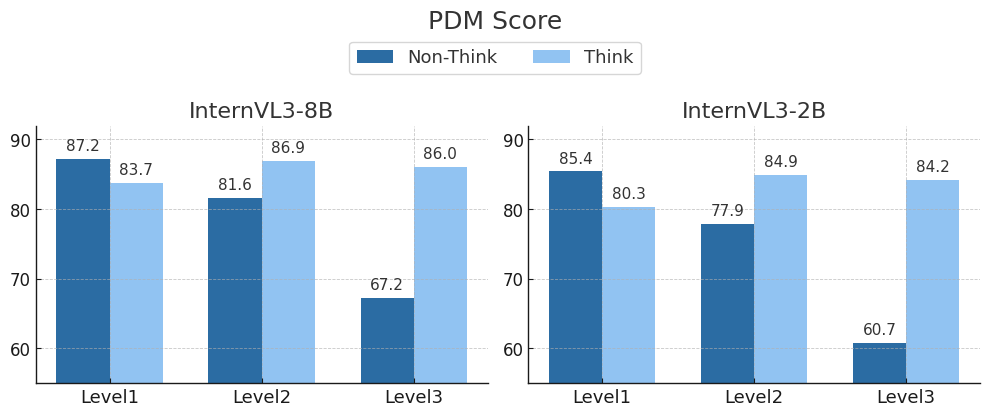}
        \caption{The performance of  CoT on InternVL3-8B/2B across scenes of varying complexity, where both denote VLM models after SFT. Scene complexity increases progressively from Level 1 (simple) to Level 3 (challenging).}
        \label{fig:fig1a}
    \end{subfigure}
    
    \begin{subfigure}{\linewidth}
        \centering
        \includegraphics[width=\linewidth]{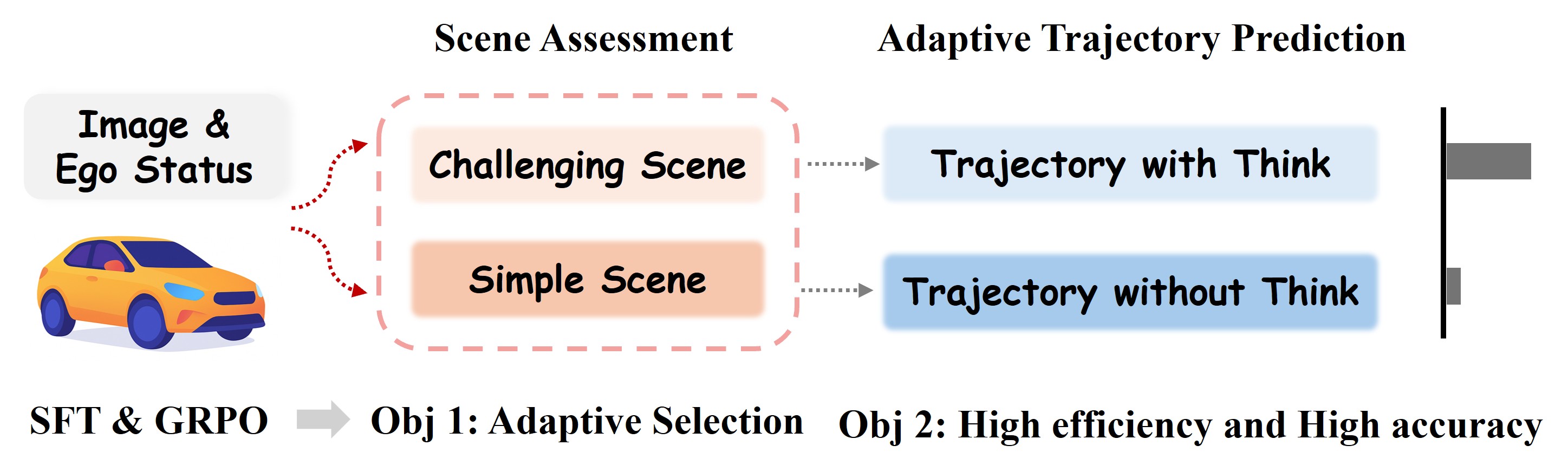}
        \caption{AdaThinkDrive learns to assess scene complexity and select response modes adaptively. It generates trajectory with Think for challenging scenes and trajectory w/o Think for simple scenes, balancing accuracy and efficiency.}
        \label{fig:fig1b}
    \end{subfigure}

    \caption{Impact and Design of Adaptive Reasoning in Trajectory Prediction.}
    \label{fig:adaptive_think_combined}
    \vspace{-1.5em}
\end{figure}

In recent years, autonomous driving systems have been shifting from traditional modular pipelines to end-to-end architectures. Although modular approaches offer engineering flexibility, they suffer from information loss across components, resulting in cumulative errors and limited generalization in complex and long-tail scenarios ~\cite{chen2024end,jiang2025survey}. End-to-end methods mitigate this by jointly optimizing perception, prediction, and planning within a unified model \cite{hu2023planning,jiang2023vad}, but their reliance on limited supervised data still constrains robustness. To address this, recent research has explored Vision-Language Models (VLMs)~\cite{vlmsurvey}, leveraging large-scale driving datasets for pretraining to enhance scene understanding capabilities~\cite{fu2025orion,wang2025omnidrive,hwang2024emma,xing2025openemma,qiao2025lightemma,tian2024drivevlm}. Current VLM-based methods fall into two categories: meta-action approaches \cite{jiang2024senna,jiang2025alphadrive}, which generate high-level guidance, and planning-based approaches \cite{hwang2024emma,li2025recogdrive}, which directly predict trajectories via language modeling. CoT has been increasingly adopted for the latter to produce structured outputs \cite{cui2025chain}, improving both interpretability and trajectory quality \cite{hwang2024emma}. However, its application to VLA in autonomous driving is still in its infancy.

To explore the potential of CoT, we conducted a comparative study on reasoning performance in VLA models under varying scene complexities. Specifically, the driving scenes were categorized into three complexity levels, as shown in \Cref{fig:fig1a}. We observe that for both InternVL3 8B and 2B models\cite{zhu2025internvl3}, the Non-Think model achieves better performance in simple scenarios (Level~1), whereas the Think model consistently outperforms as scene complexity increases (Levels 2 and 3). These findings reveal a critical limitation of existing CoT approaches: a tendency to over-reason in simple scenarios. While CoT reasoning provides substantial benefits in complex and challenging settings, it can introduce unnecessary cognitive steps and heightened uncertainty in simpler scenarios. These findings highlight that the optimal reasoning strategy is not universal but rather dependent on scene complexity. Consequently, enabling models to employ reasoning selectively based on scene complexity naturally becomes essential for improving both decision accuracy and inference efficiency in autonomous driving.

Following this, we propose \textbf{AdaThinkDrive}, a Vision-Language-Action (VLA) framework with a “Fast answering / Slow thinking” mechanism for end-to-end trajectory prediction (Figure~\ref{fig:fig1b}). 
We begin with a systematic analysis of the NAVSIM benchmark~\cite{dauner2024navsim}, evaluating the performance of existing methods across various scenario complexities. Motivated by this, we design a three-stage adaptive reasoning strategy that enables the model to automatically decide when to reason and when to act directly, guided by a learnable reward mechanism. In implementation, we first pretrain the model on large-scale driving data, then perform SFT using a customized dual-mode Navsim planning dataset to enable the model to generate both Think and Non-Think outputs. Finally, we adopt GRPO as the reinforcement learning algorithm and construct a reward structure that jointly considers trajectory accuracy, action rationality, and reasoning simplicity. This allows AdaThinkDrive to reach an optimal balance between planning performance and computational efficiency. We summarize our main contributions as follows:

\begin{itemize}
    \item We conduct comparative studies of CoT in VLA across varying levels of scenario complexity. By evaluating think and non-think paradigms on the NAVSIM benchmark, we reveal that \textit{over-reasoning in simple scenarios} appears to be a \textbf{key limitation} of existing CoT approaches, highlighting the need for adaptive reasoning strategies.

    \item We propose \textbf{AdaThinkDrive}, an end-to-end VLA framework with a “fast answering / slow thinking” mechanism that adaptively switches between direct prediction and explicit reasoning based on scene complexity. Furthermore, we design an \textbf{Adaptive Think Reward} strategy based on GRPO to guide the model in deciding when to reason and when to act directly.

    \item On the Navsim benchmark, AdaThinkDrive achieves a PDMS of 90.3, outperforming the leading vision-only baseline by 1.7 points. Furthermore, the model demonstrates its adaptive reasoning capability by selectively employing CoT in 96\% of challenging scenarios, while defaulting to direct trajectory prediction in 84\% of simple scenarios. Additionally, this adaptive approach reduces inference time by 14\% compared to an always-think baseline, confirming the framework's ability to effectively balance high performance with computational efficiency.
\end{itemize}

\section{Related Work}
\subsection{VLA for Autonomous Driving}
Vision-Language Models (VLMs) integrate visual and textual inputs for driving perception, planning, and decision-making. Existing approaches use scene understanding and high-level reasoning to guide downstream planners~\cite{tian2024drivevlm,jiang2024senna,jiang2025alphadrive,marcu2024lingoqa}, or directly predict trajectories from sensory inputs~\cite{fu2025orion,hwang2024emma,wang2025omnidrive,xing2025openemma,qiao2025lightemma,zhao2025sce2drivex,liu2025reasonplan}. Intermediate Chain-of-Thought (CoT) reasoning supports interpretable planning, as explored by EMMA~\cite{hwang2024emma}, ReasonPlan~\cite{liu2025reasonplan}, and Sce2DriveX~\cite{zhao2025sce2drivex}. ReCogDrive~\cite{li2025recogdrive} couples VLM-based driving cognition with diffusion planning and refines the planner using DiffGRPO. Our analysis instead examines when reasoning is useful: CoT primarily benefits complex scenarios and may offer little benefit in simpler scenes.

\subsection{Efficient Reasoning Models}
Long CoT in models such as DeepSeek~\cite{guo2025deepseek} improves reasoning but increases inference cost. AdaptThink~\cite{zhang2025adaptthink} studies when direct answers or explicit reasoning are preferable. Reinforcement-learning approaches address this trade-off through concise reasoning~\cite{fatemi2025concise,yi2025shorterbetter}, dynamic early stopping~\cite{hou2025thinkprune}, or on-demand reasoning conditioned on task complexity~\cite{fang2025thinkless,tu2025learning,zhang2025adaptthink,lou2025adacot}. HiPO~\cite{deng2025hipo} learns Think-on/Think-off selection using paired responses and hybrid reinforcement-learning rewards to balance accuracy and efficiency. Supervised and reinforcement post-training also support reasoning and agentic execution~\cite{zhan2025katv1,zhan2025katcoder}. In autonomous driving, simple scenes may require little reasoning, whereas complex interactions demand careful analysis before trajectory generation. AdaThinkDrive learns to switch between fast responses and slow thinking according to driving-scene complexity.

\begin{figure*}[htb]
    \centering
    \includegraphics[width=\textwidth]{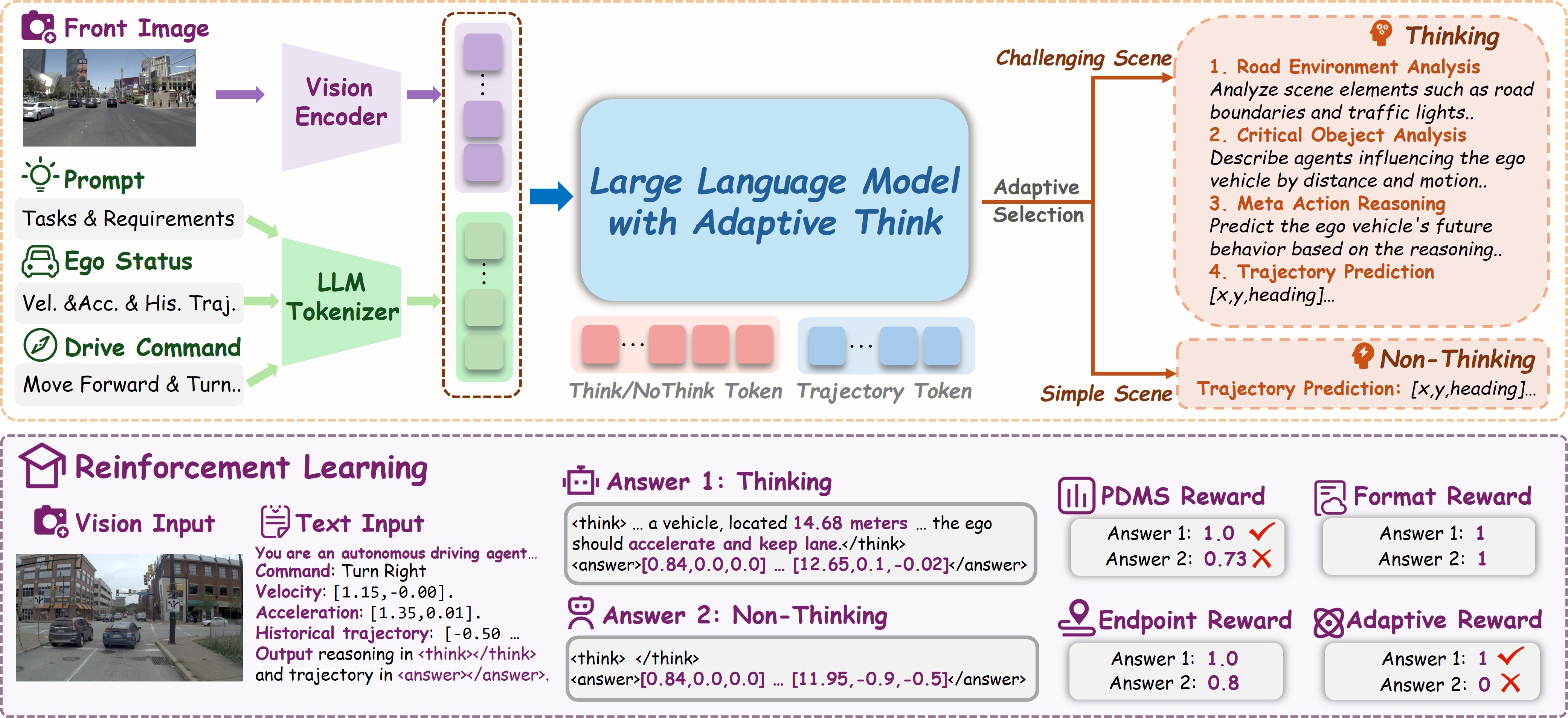}
    \caption{We present \textbf{AdaThinkDrive}, an end-to-end autonomous driving framework that adaptively selects between "Thinking" and "Non-Thinking" modes depending on scene complexity.  
    Given vision and text inputs, the VLM dynamically determines its output mode through an adaptive reasoning mechanism. During the reinforcement learning of the three-stage training process, multiple reward including PDMS, format, and endpoint are combined with the proposed Adaptive Think Reward.}
    \label{fig:fig2overview}
    \vspace{-1.5em}
\end{figure*}

\section{Methods}

In this section, we demonstrate the design of our proposed AdaThinkDrive, which includes: (1) Data preparation, including pre-training data and hybrid SFT data. (2) Processing of a two-stage Supervised fine-tuning module that provides effective initialization; (3) Adaptive complex-aware thinking via a reinforcement learning strategy that boosts the efficiency and accuracy of model. The whole framework is demonstrated in Figure~\ref{fig:fig2overview}.

\subsection{Problem Formulation}
\label{prob_def}

The input query $q$ includes a front-view image $q_{\text{cam}}$, high-level navigation commands $q_{\text{com}}$ (e.g., \textit{Move Forward, Turn Left, Turn Right}), ego state information $q_{\text{ego}}$ (e.g., \textit{velocity and acceleration}), and the historical trajectory of the last three frames $q_{\text{his}}=\{h_{t-3},h_{t-2},h_{t-1}\}$. AdaThinkDrive operates with two reasoning modes $\mathcal{M}=\{\text{Thinking},\text{Non-Thinking}\}$. Given a query $q$, the model jointly determines a reasoning mode $m\in\mathcal{M}$ and an answer $o\in\mathcal{O}$ according to the joint distribution
\begin{equation}
\vspace{-0.2em}
\mathcal{P}(m,o\,|\,q)=\mathcal{P}(m\,|\,q)\,\mathcal{P}(o\,|\,q,m).
\vspace{-0.2em}
\end{equation}
For each query $q$, the selected mode maximizes the expected task-specific utility $\mathcal{U}(q,o)$:
\begin{equation}
\vspace{-0.2em}
m(q)=\arg\max_{m\in\mathcal{M}}~\mathbb{E}_{o\sim\mathcal{P}(o\,|\,q,m)}\!\left[\mathcal{U}(q,o)\right].
\vspace{-0.2em}
\label{Eq.m*}
\end{equation}
The overall objective is to learn a policy $\pi:\mathcal{P}\to\mathcal{M}$ that selects a mode for each query so as to maximize the expected utility over query distributions $\Theta=\{(\mathcal{D}_i,\mathcal{U}_i)\}_{i=1}^N$:
\begin{equation}
\vspace{-0.2em}
\max_{\pi}\;\frac{1}{N}\sum_{i=1}^{N}
\mathbb{E}_{q\sim\mathcal{D}_i}\!\left[
\mathbb{E}_{o\sim\mathcal{P}(o\,|\,q,\pi(q))}\!\left[\mathcal{U}_i(q,o)\right]
\right].
\vspace{-0.2em}
\label{Eq.pi}
\end{equation}

\subsection{Data Preparation}

To equip the model with foundational driving knowledge and an understanding of when CoT reasoning may be beneficial, we first perform a data preparation stage as follows. 

\paragraph{Pre-training Data} To adapt general VLMs into autonomous driving, we assembled a diverse collection of open-source driving QA datasets, including DriveLM~\cite{sima2024drivelm}, LingoQA~\cite{marcu2024lingoqa}, ImpromptuVLA~\cite{chi2025impromptu}, NuScenes-QA \cite{qian2024nuscenes} NuInstruct \cite{ding2024holistic}, and OminiDrive \cite{wang2025omnidrive}. In addition, we constructed a multi-turn Q\&A reasoning dataset for NAVSIM following CoT paradigm during SFT stage, including road boundary estimation, critical object identification, ego action prediction and related scene understanding subtasks.

\begin{figure}[t!]
    \centering
    \begin{subfigure}{0.48\linewidth}
        \centering
        \includegraphics[width=\linewidth]{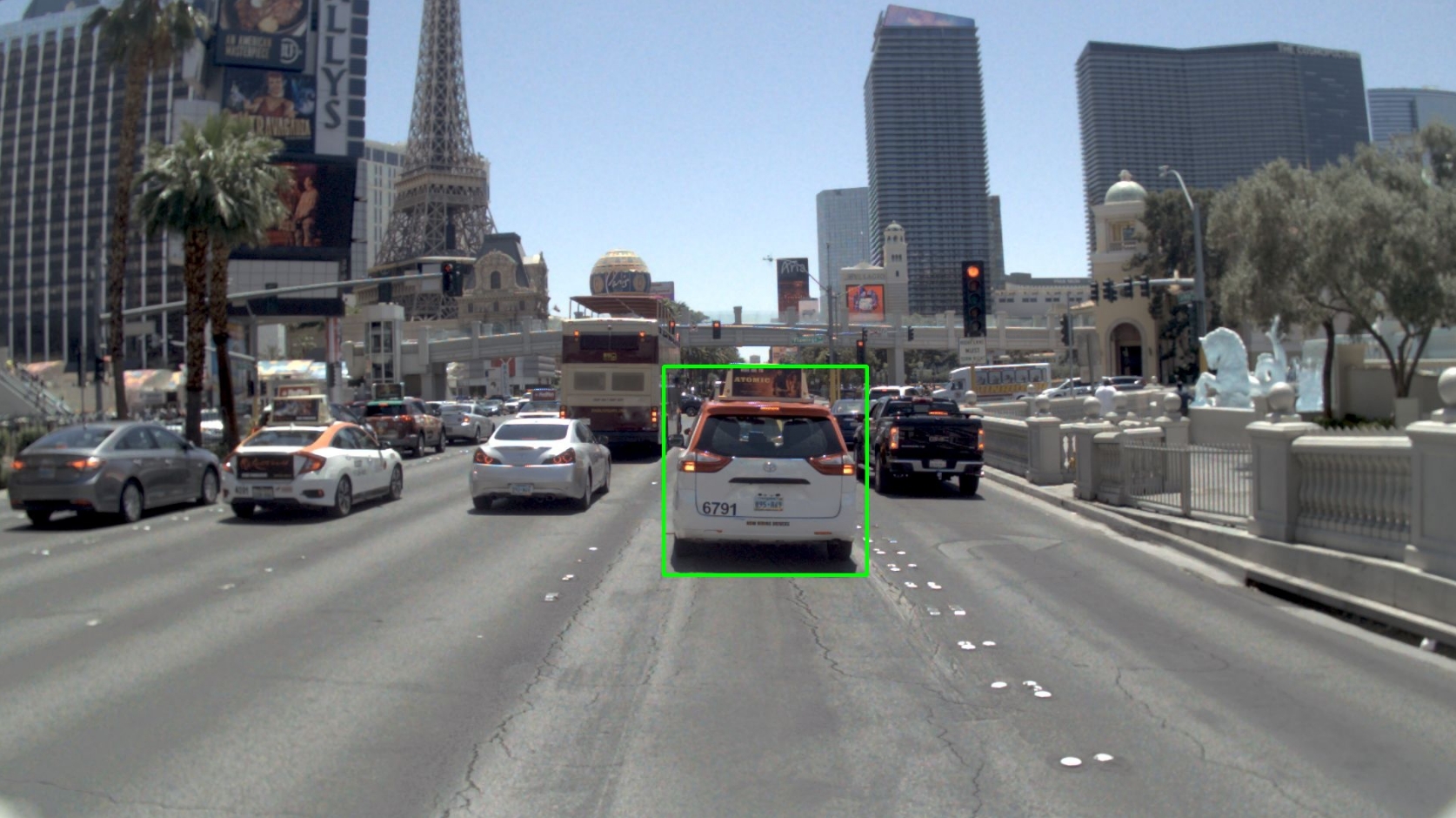}
        \caption{CIPO-1}
        \label{fig:cipo1}
    \end{subfigure}
    \hfill 
    \begin{subfigure}{0.48\linewidth}
        \centering
        \includegraphics[width=\linewidth]{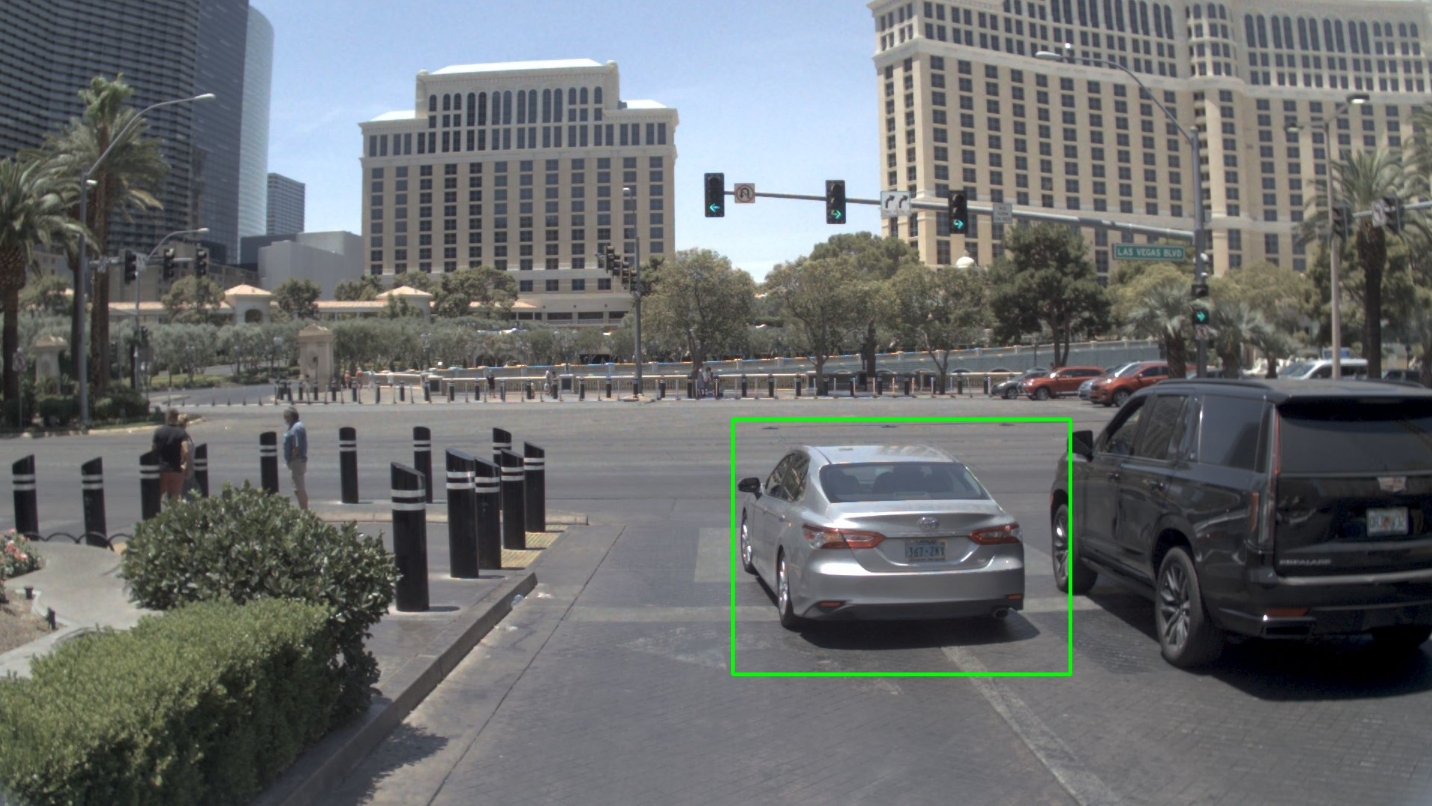}
        \caption{CIPO-2}
        \label{fig:cipo2}
    \end{subfigure}

    \vspace{1em} 

    \begin{subfigure}{0.48\linewidth}
        \centering
        \includegraphics[width=\linewidth]{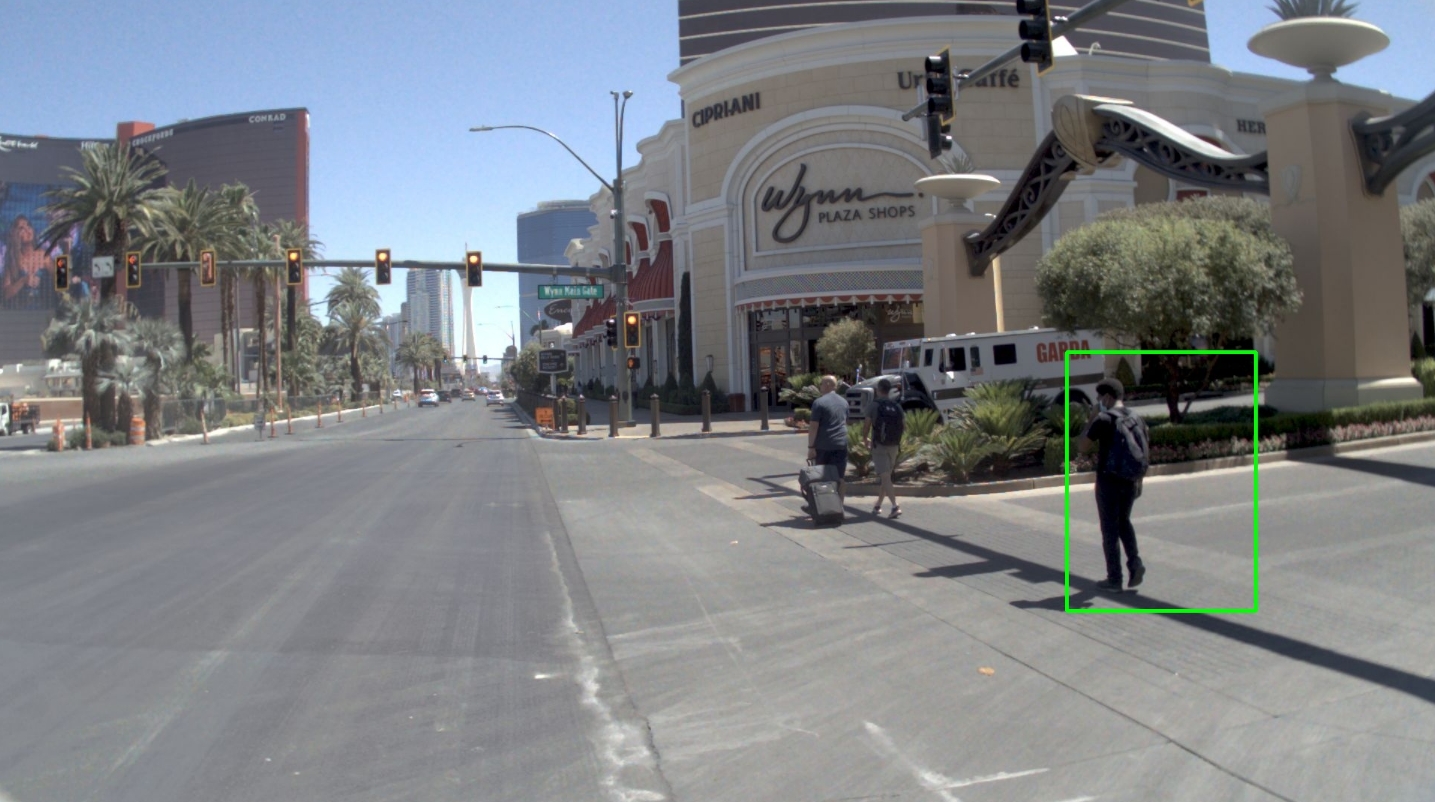}
        \caption{Motion Interaction}
        \label{fig:motion}
    \end{subfigure}
    \hfill
    \begin{subfigure}{0.48\linewidth}
        \centering
        \includegraphics[width=\linewidth]{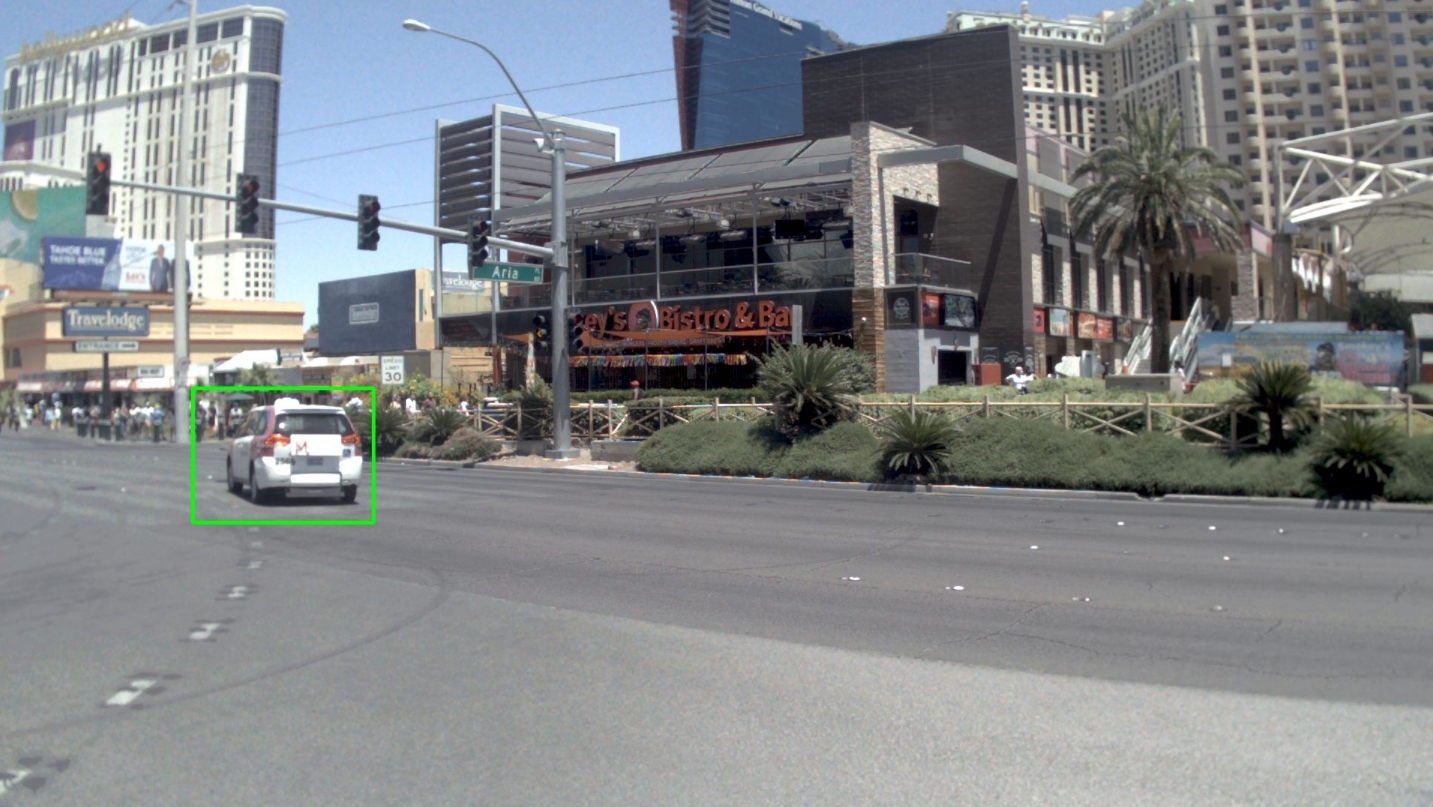}
        \caption{Motion Interaction}
        \label{fig:motion2}
    \end{subfigure}
    
    \vspace{0.5em}
    \caption{Visualization of dynamic agents for Think-style CoT supervision. (a)-(b) depict CIPO-1 agents (occupying the ego lane) and CIPO-2 agents (likely to merge), while (c)-(d) show Motion Interaction cases where agents’ future trajectories intersect with the ego vehicle’s trajectory.}
    \label{fig:cot}
    \vspace{-1.5em}
\end{figure}

\paragraph{Hybrid SFT Data} The SFT dataset consists of both reasoning-intensive and direct-answer examples. Reasoning data is generated by an auxiliary model combined with rule-based methods, ensuring high-quality reasoning traces. For detailed scene descriptions like traffic light states and weather conditions, we automatically generate fine-grained annotations with Qwen2.5-VL-72B. Furthermore, to capture interactive scene dynamics, we identify dynamic agents expected to interact with the ego vehicle and categorize them into three types: Closest In-Path Object (CIPO-1) for those in the ego lane; CIPO-2 for those likely to merge, determined by lane geometry and relative location; and Motion Interaction for those with future trajectories predicted to intersect the ego trajectory, as illustrated in Figure~\ref{fig:cot}. For static elements such as road boundaries, we utilize the NAVSIM map to reconstruct lane topology and find critical boundary features along the ego’s future path.

Furthermore, for each query $q$ in the dataset, we generate both a Think-style response $\{q, o^{Think}\}$, which retains the full reasoning process $<$think$>$$reasoning$$<$/think$>$, and a Non-Think-style response $\{q, o^{Non\text{-}Think}\}$, which omits explicit reasoning but maintains structural consistency. For all cases, we directly fill the trajectory into $<$answer$>$$trajectory$$<$/answer$>$. Collectively, we denote the resulting supervised dataset as 
$\mathcal{D}^{\text{SFT}}=\{\{q, o^{Think}\}, \{q, o^{Non\text{-}Think}\}\}_{q\in\mathcal{Q}}$. 
This SFT data serves as a ``warm-up'', equipping the model with foundational capability to distinguish between two response styles. The demo pipeline can be found in \Cref{fig:fig2overview}.

\paragraph{Scene Categorization} To support adaptive reasoning, we categorize the NAVSIM training and validation sets into three levels of increasing complexity: Level 1, Level 2, and Level 3. This division follows the same criteria used in constructing the Think-style and non-Think-style datasets, which are based on two factors: whether the ego vehicle is near road boundaries and whether critical objects are present that may influence driving decisions. Level 1 includes scenes with neither condition, Level 2 includes scenes with only one, and Level 3 includes scenes with both. We define the dataset as $\mathcal{D}=\{\mathcal{D}^{+},\mathcal{D}^{-}\}$, where $\mathcal{D}^{-}$ corresponds to Level~1 and $\mathcal{D}^{+}$ corresponds to Level~2 and~3. These two categories serve as auxiliary labels to provide a principled initialization for downstream reinforcement learning.

\subsection{Two-Stage Supervised Fine-tuning Processing}
To build a model with driving knowledge and trajectory planning capabilities, we perform a two-stage fine-tuning procedure. The first stage injects general driving knowledge, while the second focuses on equipping models with the ability to adhere to trajectory generation and output formats.

In the first stage, the model is pretrained on a large corpus of driving-related Q\&A pairs, enhancing its understanding of driving domain cognition. This phase addresses tasks like understanding the drivable area, object localization, and traffic semantics.

Subsequently, the second stage introduces the trajectory prediction task. For each query $q = (q_{\text{cam}}, q_{\text{com}}, q_{\text{ego}}, q_{\text{his}})$, two outputs are generated: $o^{\text{Thinking}}$, which includes reasoning, and $o^{\text{Non-Thinking}}$, which contains only the final trajectory.

During fine-tuning, the model is supervised on both outputs, aiming to maximize the conditional likelihood:
\begin{equation}
\vspace{-0.2em}
\mathcal{L}_{\text{SFT}} = 
\mathbb{E}_{(q, o) \sim \mathcal{D^{SFT}}} \big[ -\log \pi_\theta(o \mid q) \big].
\vspace{-0.2em}
\end{equation}

This training strategy enables the model to learn both \textit{Thinking} and \textit{Non-Thinking} reasoning modes under a unified interface, while remaining unbiased toward either style. As a result, the model can generate both response types for any query $q$, enabling adaptive reasoning in the GRPO phase.

\subsection{ Adaptive Thinking via Reinforcement Learning}
After SFT processing, AdaThinkDrive acquires the initial ability to support two different reasoning modes on the same query $q$ without collapsing. However, our goal is to enable the policy model to adaptively select the most appropriate reasoning mode $m(q)$ to improve its efficiency. To this end, we introduce a reinforcement learning phase in addition to supervised learning to explicitly teach the model how to adaptively select between reasoning modes, while simultaneously improving the model’s planning ability.

To enable the model to learn not only how but also when to reason, and to effectively leverage the reasoning process for accurate trajectory prediction, we design four complementary reward components: PDMS Reward, Format Reward, Endpoint Reward, and Adaptive Think Reward.

\subsubsection{PDMS Reward} Evaluation metric \textit{Predictive Driver Model Score (PDMS)} \cite{dauner2024navsim} for predicted trajectory is used for trajectory reward $\mathcal{R}_{\text{traj}}$, which is a Continuous value from 0 to 1.

\subsubsection{Format Reward} This reward $\mathcal{R}_{fmt}$ enforces compliance with the prescribed output format, covering both the correct use of $<$think$>$...$<$/think$>$ and $<$answer$>$...$<$/answer$>$ tags and the standardized representation of predicted trajectories. It provides discrete feedback for violations in either component, thereby ensuring consistent structural and content formatting.

\subsubsection{Endpoint Reward} To encourage accurate alignment between predicted and ground-truth trajectory endpoints, we adopt a piecewise reward $\mathcal{R}_{endpoint}$ based on the L1 distance of the final point. A full score of 1.0 is given when the deviation is below 2 meters, decreasing stepwise to 0.8 ($<$4~m), 0.6 ($<$6~m), 0.4 ($<$10~m), 0.2 ($<$~15m), and 0.0 otherwise. This design penalizes large errors while remaining sensitive to small deviations near the trajectory endpoint.

\begin{figure}[htbp]
    \centering
    \includegraphics[width=\linewidth]{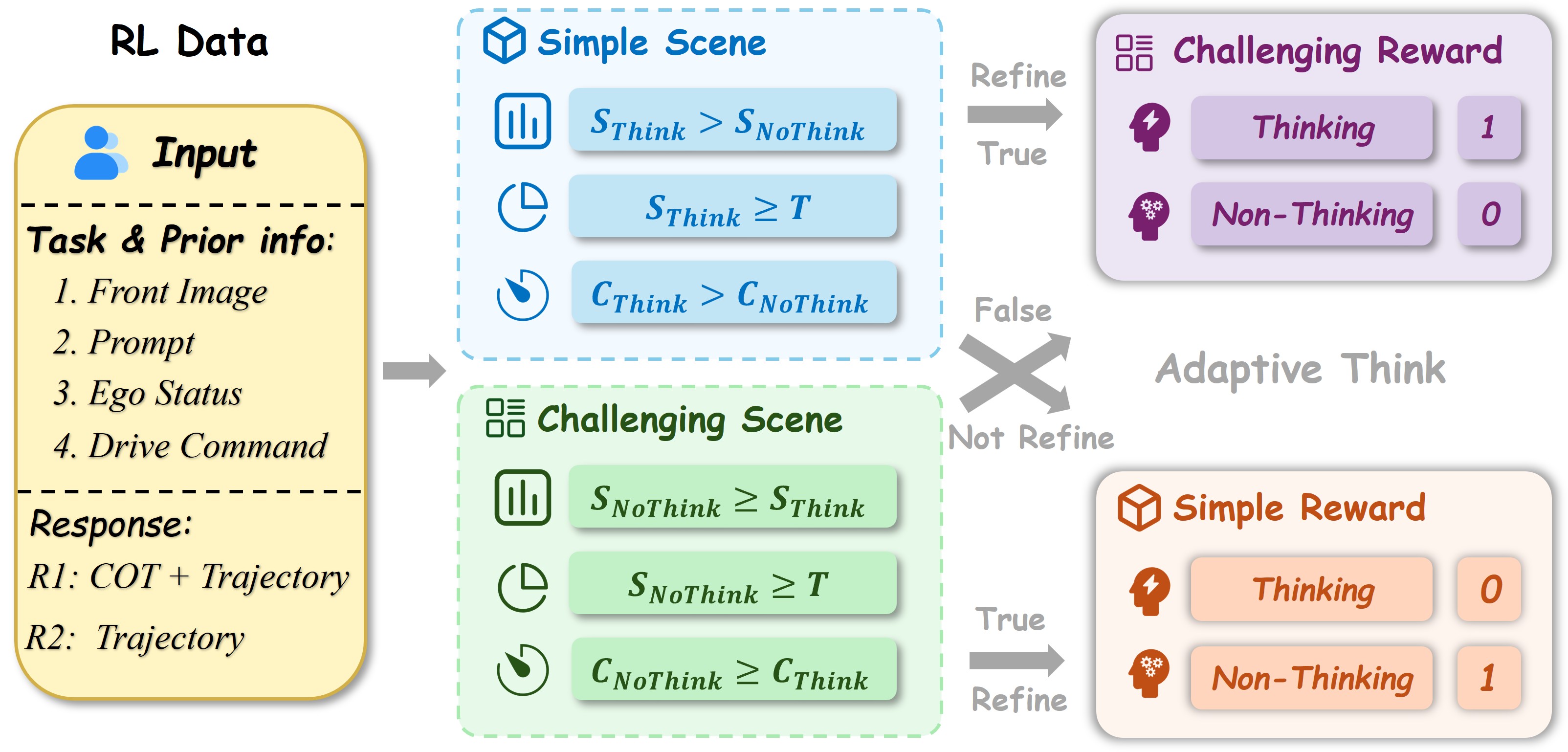}
    \caption{Adaptive Think Reward: A Dynamic Reasoning Control Strategy. This reward adjusts the model’s reasoning behavior by identifying misclassified scenes. When scene-specific conditions are satisfied, it assigns rewards to either Thinking or Non-thinking responses accordingly.}
    \label{fig:adaptive_think_reward}
    \vspace{-1.0em}
\end{figure}

\subsubsection{Adaptive Think Reward} During reinforcement learning training, the policy model’s capability evolves dynamically. The same scenarios may be perceived as simple or challenging at different stages of training. To address for this, \textit{Adaptive Think Reward} $\mathcal{R}_{adaptive}$ is designed to teach the model when to think, preventing over-reliance on static, manually defined scene tags$\{\mathcal{D^{+}}, \mathcal{D^{-}}\}$. These manual tags serve as an initial basis for reasoning, helping the model avoid ``collapse", where it only outputs Thinking or Non-Thinking without adapting to the reasoning needs of the situation. The dynamic adjustment mechanism allows the model to correct scene tags based on actual reasoning needs, gradually learning adaptive thinking and improving both decision-making efficiency and prediction accuracy.

Adaptive Think Reward guides the model to adjust reasoning behavior dynamically using multiple rollouts. The detailed process is illustrated in Figure~\ref{fig:adaptive_think_reward} and Algorithm~\ref{alg:adaptive_think_algorithm}.

\begin{algorithm}
\caption{Adaptive Think Reward}
\label{alg:adaptive_think_algorithm}
\begin{algorithmic}[1]
\small
\Require
\Statex $S_{\text{Think}}$: Average PDMS of Thinking rollouts  
\Statex $S_{\text{Nothink}}$: Average PDMS of Non-thinking rollouts  
\Statex $C_{\text{Think}}$: Count of Thinking rollouts  
\Statex $C_{\text{Nothink}}$: Count of Non-thinking rollouts  
\Statex $T$: Confidence threshold (default: 0.9)  
\Statex $D$: Scene complexity label (0: Simple, 1: Challenging)  

\Ensure
\Statex $\text{Reward}_{\text{Thinking}}, \text{Reward}_{\text{Non-Thinking}}$

\Statex \textbf{Case $D = 0$ (Simple scene):}
\If{$S_{\text{Think}} > S_{\text{Nothink}}$ \& $S_{\text{Think}} > T$ \& $C_{\text{Think}} > C_{\text{Nothink}}$}
    \Statex \hspace{1em} \textit{// Corrected to Challenging}
    \State $\text{Reward}_{\text{Thinking}} \gets 1$, $\text{Reward}_{\text{Non-Thinking}} \gets 0$
\Else \hspace{1em} \textit{// Maintained as Simple}
    \State $\text{Reward}_{\text{Thinking}} \gets 0$, $\text{Reward}_{\text{Non-Thinking}} \gets 1$
\EndIf

\Statex \textbf{Case $D = 1$ (Challenging scene):}
\If{$S_{\text{Nothink}} > S_{\text{Think}}$ \& $S_{\text{Nothink}} > T$ \& $C_{\text{Nothink}} > C_{\text{Think}}$}
    \Statex \hspace{1em} \textit{// Corrected to Simple}
    \State $\text{Reward}_{\text{Thinking}} \gets 0$, $\text{Reward}_{\text{Non-Thinking}} \gets 1$
\Else \hspace{1em} \textit{// Maintained as Challenging}
    \State $\text{Reward}_{\text{Thinking}} \gets 1$, $\text{Reward}_{\text{Non-Thinking}} \gets 0$
\EndIf

\State \Return $\text{Reward}_{\text{Thinking}}, \text{Reward}_{\text{Non-Thinking}}$
\end{algorithmic}
\end{algorithm}

The overall reward in the reinforcement learning process is computed by integrating four specifically designed reward components, which present as follows:
\begin{equation}
\vspace{-0.2em}
    \mathcal{R}(q, a) = \mathcal{R}_{traj} + \mathcal{R}_{fmt} + \mathcal{R}_{endpoint} + \mathcal{R}_{adaptive}.
\vspace{-0.2em}
\end{equation}

We utilize GRPO~\cite{shao2024deepseekmath} as the training algorithm. For each query $q$, a set of candidate outputs $\{o_1, o_2, \dots, o_G\}$ is sampled from the old policy $\pi_{\text{old}}$, and the current policy $\pi_\theta$ is optimized based on their reward signals. To ensure stable training and avoid drastic policy shifts, GRPO incorporates truncated importance weights and a KL divergence regularization term: the former suppresses excessive policy updates, while the latter constrains the current policy from deviating too far from the reference policy $\pi_{\text{ref}}$. The final optimization objective for GRPO is defined as follows:
\begin{align}
    \mathcal{J}(\theta) &= \mathbb{E}_{q, \{o_i\} \sim \pi_{\theta_{old}}} \left[ \frac{1}{G} \sum_{i=1}^G \mathcal{J}_i - \beta \mathbb{D}_{KL}(\pi_{\theta} || \pi_{ref}) \right],\\
    \mathcal{J}_i &= \min \big( c_i A_i, \text{clip}\left(c_i, 1 - \epsilon, 1 + \epsilon\right) A_i \big).
\end{align}

\noindent where $c_i = \frac{\pi_\theta(o_i |q)}{\pi_{\theta_{old}}(o_i |q)}$, $\epsilon$ and $\beta$ are hyper-parameters, and the relative advantage $A_i$ is calculated based on the normalized reward difference of a set of candidate outputs. 

Through reinforcement learning, the policy model can develop adaptive reasoning strategies that adjust dynamically to the scenarios of different complexity.

\section{Experiment}

\begin{table*}[t]
\small\setlength{\tabcolsep}{4.5pt}\renewcommand{\arraystretch}{1}
\centering
\caption{The Performance Comparison on NAVSIM using Closed-Loop Metrics.}
\begin{tabular}{l|c|c|rrrrr|r}
\toprule
Method & Image & Lidar & NC$\uparrow$ & DAC$\uparrow$ & TTC$\uparrow$ & CF$\uparrow$ & EP$\uparrow$ & PDMS$\uparrow$ \\
\midrule
Constant Velocity & & & 68.0 & 57.8 & 50.0 & \textbf{100} & 19.4 & 20.6 \\
Ego Status MLP & & & 93.0 & 77.3 & 83.6 & \textbf{100} & 62.8 & 65.6 \\
\midrule
UniAD \cite{hu2023planning} & \checkmark & & 97.8 & 91.9 & 92.9 & \textbf{100} & 78.8 & 83.4 \\
LFT \cite{chitta2022transfuser} & \checkmark & & 97.4 & 92.8 & 92.4 & \textbf{100} & 79.0 & 83.8 \\
TransFuser \cite{chitta2022transfuser} & \checkmark & \checkmark & 97.7 & 92.8 & 92.8 & \textbf{100} & 84.0 & 84.0 \\
PARA-Drive \cite{weng2024drive} & \checkmark & & 97.9 & 92.4 & 93.0 & 99.8 & 79.3 & 84.0 \\
DRAMA \cite{yuan2024drama} & \checkmark & \checkmark & 98.0 & 93.1 & 94.8 & \textbf{100} & 80.1 & 85.5 \\
Hydra-MDP-$\mathcal{V}_{8192}$-W-EP \cite{li2024hydra} & \checkmark & \checkmark & 98.3 & 96.0 & 94.6 & 100 & 78.7 & 86.5 \\
DiffusionDrive \cite{liao2025diffusiondrive} & \checkmark & \checkmark & 98.2 & 96.2 & 94.7 & \textbf{100} & 82.2 & 88.1 \\
WoTE \cite{li2025end} & \checkmark & \checkmark & \textbf{98.5} & 96.8 & 94.9 & 99.9 & 81.9 & 88.3 \\
Hydra-NeXt \cite{li2025hydra} & \checkmark & & 98.1 & 97.7 & 94.6 & \textbf{100} & 81.8 & 88.6\\
GoalFlow \cite{xing2025goalflow} & \checkmark & \checkmark & 98.4 & \textbf{98.3} & 94.6 & \textbf{100} & \textbf{85.0} & \textbf{90.3} \\
\midrule
\rowcolor{gray!15}
\textbf{AdaThinkDrive} (Ours) & \checkmark & & 98.4 & 97.8 & \textbf{95.2} & \textbf{100} & 84.4 & \textbf{90.3} \\
\rowcolor{gray!15}
\textbf{AdaThinkDrive} (Best-of-N) & \checkmark & & 99.1 & 98.8 & 97.2 & 100 & 87.9 & 93.0 \\
\bottomrule
\end{tabular}
\label{table:main_table}

\end{table*}

\begin{table}[t]
\centering
\footnotesize 
\setlength{\tabcolsep}{2.5pt}
\caption{Comparison of Think and Non-Think Models in SFT and RL (InternVL3-8B). We compare the performance of SFT and RL models in Thinking (w) and Non-Thinking (w/o) modes. "Ours" refers to adaptive think mode.}
\begin{tabular}{@{}c|c|ccccc|c@{}}
\toprule
Model & Mode & NC$\uparrow$ & DAC$\uparrow$ & TTC$\uparrow$ & CF$\uparrow$ & EP$\uparrow$ & PDMS$\uparrow$ \\
\midrule
Non-Think SFT & w/o & 98.1 & 92.1 & 94.0 & 100 & 77.1 & 83.3 \\
Think SFT & w & 98.5 & 94.4 & 94.9 & 100 & 79.9 & 86.2 \\
\midrule
Non-Think RL & w/o & 98.2 & 96.1 & 94.3 & 100 & 83.5 & 88.3 \\
Think RL & w & 98.2 & 96.4 & 94.6 & 100 & 84.2 & 88.9 \\
\textbf{AdaThinkDrive} & Ours & 98.4 & \textbf{97.8} & \textbf{95.2} & \textbf{100} & \textbf{84.4} & \textbf{90.3} \\
\bottomrule
\end{tabular}
\label{table:compa_think_and_nothink}

\end{table}

\begin{table}[t]
\centering
\footnotesize 
\setlength{\tabcolsep}{4pt} 
\caption{Comparison of inference time and PDMS among the Non-Think Model, Think Model, and AdaThinkDrive. Inference time denotes the average time to predict 4-second trajectories on the NAVSIM Test dataset.}
\begin{tabular}{c|c|c}
\toprule
Method & Infer Time$\downarrow$ & PDMS$\uparrow$ \\
\midrule
Non-Think RL & \textbf{0.68} & 88.3 \\
Think RL  & 0.86 & 88.9 \\
\midrule
\textbf{AdaThinkDrive} & 0.74 & \textbf{90.3} \\
\bottomrule
\end{tabular}
\label{table:speed}

\end{table}

\begin{table}[t]
\centering
\footnotesize 
\setlength{\tabcolsep}{1.5pt}
\caption{Comparison of Think and Non-Think RL Models in Simple (Level 1) and Challenging (Level 3) Scenarios.}
\begin{tabular}{l|c|ccccc|c}
\toprule
\textbf{Model} & Level & NC$\uparrow$ & DAC$\uparrow$ & TTC$\uparrow$ & CF$\uparrow$ & EP$\uparrow$ & PDMS$\uparrow$ \\
\midrule
\multirow{2}{*}{\shortstack[l]{Non-Think RL\\\textbf{AdaThinkDrive}}} & \multirow{2}{*}{Level 1} 
& 98.5 & 96.3 & 95.0 & 100 & 83.0 & 88.5 \\
& & \textbf{98.8} & \textbf{98.1} & \textbf{96.1} & \textbf{100} & \textbf{84.5} & \textbf{90.7} \\
\midrule
\multirow{2}{*}{\shortstack[l]{Think RL\\\textbf{AdaThinkDrive}}} & \multirow{2}{*}{Level 3} 
& 98.7 & 92.9 & 95.4 & 100 & 84.7 & 87.8 \\
& & \textbf{99.4} & \textbf{94.6} & \textbf{96.3} & \textbf{100} & \textbf{86.6} & \textbf{89.8} \\
\bottomrule
\end{tabular}
\label{tab:reasoning_results}

\par\vspace{8pt}
\begingroup
\captionsetup{type=figure}
    \centering
    \includegraphics[width=\linewidth]{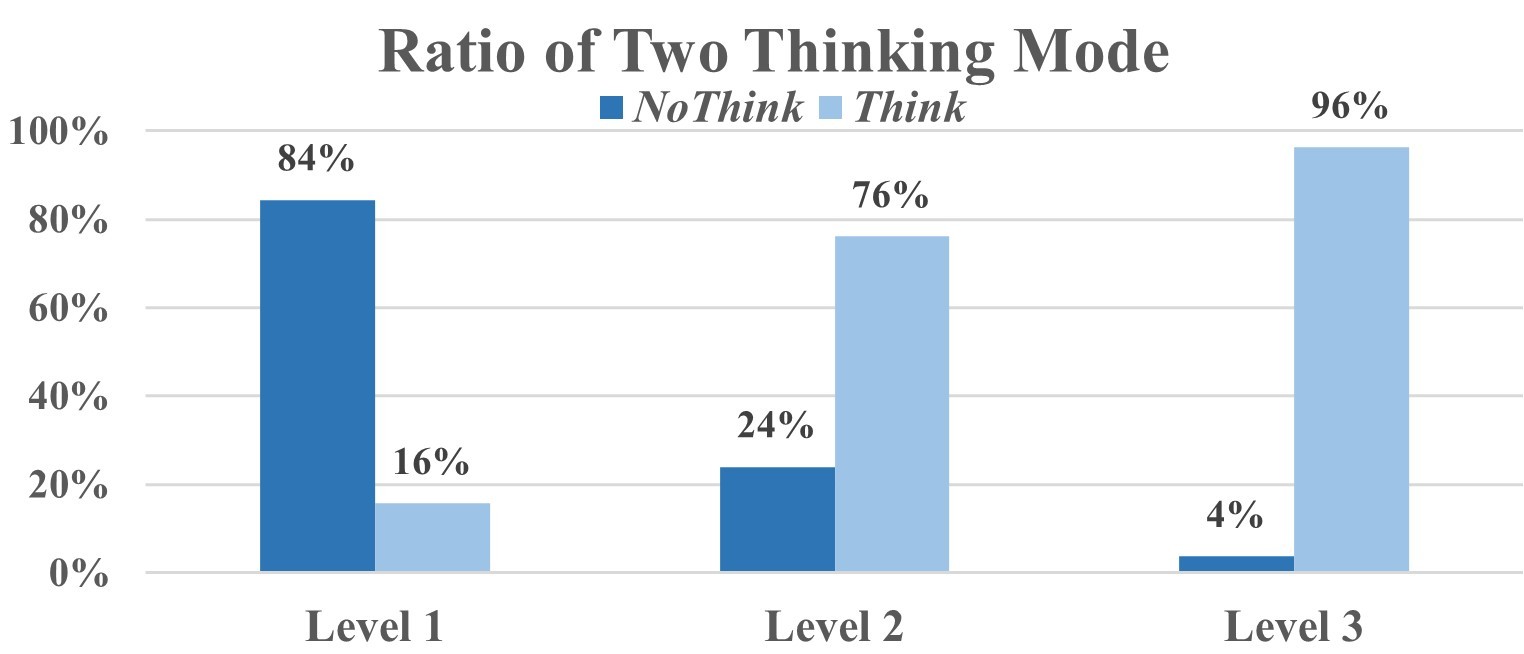}
    \caption{The ratio of Think vs. Non-Think choices by AdaThinkDrive across different NAVSIM Test dataset levels. Scene complexity increases progressively from Level 1 (simple) to Level 3 (challenging).}
    \label{fig:ratio}
\endgroup
\end{table}

\subsection{Dataset and Metric}
\subsubsection{Dataset} We conduct comprehensive experiments and evaluations on NAVSIM, a planning-oriented autonomous driving dataset built on the OpenScene platform. In addition to the reasoning data collected from NAVSIM, we further leverage several open-source datasets, such as DriveLM, ImpromptuVLA, and LingoQA, by reformatting their visual VQA pairs to better support CoT reasoning.

\subsubsection{Metric} The NAVSIM benchmark provides a non-reactive simulation environment and employs the Predictive Driver Model Score (PDMS) as closed-loop planning metric:
\begin{equation}
PDMS = NC \times DAC \times\! \left( \frac{5\times EP + 5\times TTC + 2\times C}{12} \right).
\end{equation}
where PDMS integrates five sub-metrics: No At-Fault Collision (NC), Drivable Area Compliance (DAC), Time-to-Collision (TTC), Comfort (Comf.), and Ego Progress (EP) to produce a comprehensive closed-loop planning score. 
\subsection{Implementation Details}
We use InternVL3-8B \cite{zhu2025internvl3} as the base model. The training consists of three stages. In the first stage, we fine-tune on a large-scale driving knowledge dataset for 2 epochs with a learning rate of $1 \times 10^{-5}$ and batch size 1. The second stage fine-tunes on a curated Navsim planning dataset with Think and Non-Think annotations for 2 epochs with a learning rate of $4 \times 10^{-5}$ and batch size 2. The third stage applies reinforcement learning for 2 epochs with a learning rate of $2 \times 10^{-6}$, batch size 4, and 64 NVIDIA H20 GPUs. The threshold $T$ for Adaptive Think reward is set to 0.9.

\subsection{Performance Comparison}
\subsubsection{AdaThinkDrive Performance} Table~\ref{table:main_table} presents the performance comparison between AdaThinkDrive and current leading methods on the NAVSIM benchmark. Under the vision-only setting, AdaThinkDrive achieves a PDMS of 90.3, establishing a new state-of-the-art (SOTA). Compared with the previous best vision-only method, Hydra-NeXt, AdaThinkDrive improves PDMS by 1.7, demonstrating significant advances in modeling capability and trajectory prediction accuracy. Moreover, despite relying solely on vision input, AdaThinkDrive performs comparably to the multi-modal approach GoalFlow, further validating the effectiveness of its adaptive reasoning mechanism and its strong generalization ability in complex driving scenarios. Finally, in best-of-N planning, we use the Navsim reference trajectory evaluator to select the optimal trajectory from four generated candidates, ultimately achieving the highest PDMS of 93.0.

\begin{figure*}[t]
    \centering
    \includegraphics[width=0.93\textwidth]{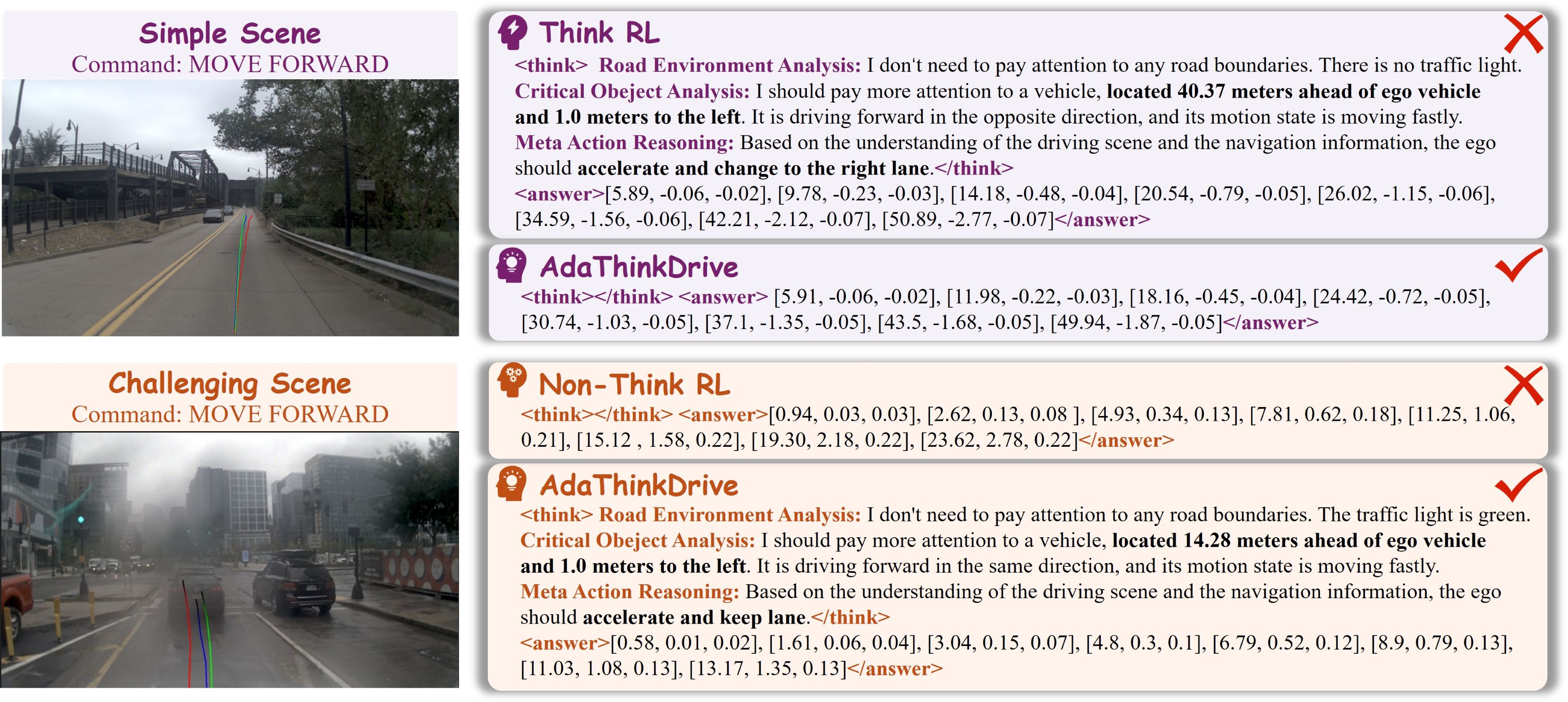}
    \caption{Qualitative comparisons of trajectory predictions on the NAVSIM dataset. We compare AdaThinkDrive with the Think RL model in the simple scenario (\textbf{top}), and with the Non-Think RL model in a challenging scenario (bottom). Each example shows the input image, driving command, reasoning content (if any), and the predicted trajectory. Green, blue, and red represent the ground-truth, AdaThinkDrive planning, and Think/Non-Think Model planning, respectively.}
    \label{fig:fig6visual}
\vspace{-1.5em}
\end{figure*}

\subsubsection{Quantitative Evaluation of Adaptive Think}

We first compare AdaThinkDrive (Table~\ref{table:compa_think_and_nothink}) against several models:
\begin{itemize}
  \item \textbf{Think/Non-Think SFT:} SFT model trained to always or never generate CoT.
  \item \textbf{Think/Non-Think RL:} RL model fine-tuned from the corresponding Think or Non-Think SFT model.
\end{itemize}

Notably, AdaThinkDrive achieves the best overall performance, outperforming the Non-Think RL and Think RL baselines by 2.0 and 1.4 PDMS, respectively. As shown in Table~\ref{tab:reasoning_results}, it achieves 2.2 higher PDMS than Non-Think RL in Level 1 (simple scenarios) and 2.0 higher than Think RL in Level 3 (challenging scenarios). These improvements highlight AdaThinkDrive’s ability to combine the advantages of both strategies: skipping reasoning in simple scenarios to enhance efficiency, while leveraging structured reasoning in challenging ones for greater accuracy. Behavior analysis across different scene complexity (Figure~\ref{fig:ratio}) further confirms that AdaThinkDrive prefers the Non-Think mode in simple scenes and increasingly adopts the Think mode in challenging ones, demonstrating its capacity for dynamic reasoning control. In addition, Table~\ref{table:speed} shows our model’s inference time is 9\% higher than Non-Think RL baseline, while achieving a notable improvement of 2.0 PDMS in accuracy. Moreover, it shows a 14\% reduction in inference time compared to Think RL baseline. Overall, these results validate the effectiveness of adaptive reasoning in balancing accuracy and efficiency across diverse driving scenarios.

\subsubsection{Qualitative Analysis of Adaptive Think} Figure~\ref{fig:fig6visual} shows qualitative comparisons between AdaThinkDrive and baselines in both simple and challenging scenarios. In the simple scenario, the Think model misclassifies a distant object as critical, leading to unnecessary reasoning and a trajectory that strays from the drivable area. AdaThinkDrive skips redundant reasoning and directly outputs a smooth, accurate trajectory. In the challenging case, the Non-Think RL model fails to assess the distance to the lead vehicle, resulting in a risky plan. In contrast, AdaThinkDrive identifies the critical object and generates a safe trajectory. These examples demonstrate its ability to adaptive reasoning to scene complexity, improving both safety and decision quality.

\subsection{Ablation Studies}

\suppressfloats[t]
\begin{table}[t]
\centering
\footnotesize 
\setlength{\tabcolsep}{3.6pt}
\caption{Ablation Study on AdaThinkDrive Components. We evaluate the effect of pre-training, supervised fine-tuning, and reinforcement learning on driving performance using NAVSIM evaluation metrics.}
\begin{tabular}{l|ccccc|c}
\toprule
Model & NC$\uparrow$ & DAC$\uparrow$ & TTC$\uparrow$ & CF$\uparrow$ & EP$\uparrow$ & PDMS$\uparrow$ \\
\midrule
SFT & 98.5 & 94.4 & 94.9 & 100 & 79.9 & 86.2 \\
Pre+SFT & 98.9 & 95.3 & 96.0 & 100 & 80.6 & 87.5 \\
Pre+SFT+RL & 98.4 & \textbf{97.8} & 95.2 & \textbf{100} & \textbf{84.4} & \textbf{90.3} \\
\bottomrule
\end{tabular}
\label{table:ablation_train}

\end{table}

\subsubsection{Alabtion study on AdaThinkDrive} 
Table~\ref{table:ablation_train} presents the ablation results of the three-stage training pipeline for AdaThinkDrive. Using only NAVSIM trajectory data for SFT, the model achieves a PDMS of 86.2. Adding pretraining on a large-scale driving QA dataset boosts the score to 87.5, an increase of 1.3. Incorporating Adaptive Think reinforcement learning and the Adaptive Think Reward further improves performance to 90.3 PDMS, a gain of 2.8. These results demonstrate that both pretraining and the adaptive reinforcement learning strategy play a crucial role in enhancing the model's understanding and reasoning capabilities.

\subsubsection{Comparison of Reward Effectiveness}
Table~\ref{table:albtion_reward} illustrates the impact of different reward combinations on PDMS. With only basic PDMS and format rewards, the model reaches 88.1. Adding the Endpoint Reward slightly raises it to 89.1. Incorporating Adaptive Think Reward further improves PDMS to 90.3, showing that adaptive reasoning is essential for improving planning efficiency and accuracy, and enhancing decision-making across diverse scenarios.

\begin{table}[t]
\centering
\footnotesize 
\setlength{\tabcolsep}{2.0pt}
\caption{Ablation Studies of Reward Designs in GRPO. We evaluate the effect of PDMS Reward (P.), Format Reward (F.), Endpoint Reward (E.), and Adaptive Think Reward (A.) on driving performance using NAVSIM evaluation metrics.}
\begin{tabular}{c| c c c|c c c c c|c}
\toprule
ID & P. \& F. & E. & A. & NC$\uparrow$ & DAC$\uparrow$ & TTC$\uparrow$ & CF$\uparrow$ & EP$\uparrow$ & PDMS$\uparrow$ \\
\midrule
1 & \checkmark & & & 96.7 & 95.2 & 90.7 & 100 & 87.8 & 88.1 \\
2 & \checkmark & \checkmark & & 98.3 & 96.6 & 94.5 & 100 & 84.4 & 89.1 \\
3 & \checkmark & \checkmark & \checkmark & \textbf{98.4} & \textbf{97.8} & \textbf{95.2} & \textbf{100} & \textbf{84.4} & \textbf{90.3} \\
\bottomrule
\end{tabular}
\label{table:albtion_reward}

\end{table}

\section{Conclusion}
In this paper, we argue that reasoning in simple scenarios often introduces computational overhead without improving decision quality. To overcome this, We introduced AdaThinkDrive, a vision-language-action framework that enables the agent to adaptively learn when to think. Our key contribution is a reinforcement learning framework guided by an Adaptive Think Reward, which aligns reasoning behavior with scene complexity. Experimental results on the NAVSIM benchmark demonstrate that AdaThinkDrive achieves SOTA performance. These findings highlight the importance of adaptive thinking for achieving both accurate and efficient decision-making in autonomous systems.

\section{ACKNOWLEDGEMENT}
This work was funded in part by the National Natural Science Foundation of China under Grant 62461160260, National Key Research and Development Program of China under grant 2023YFE0205800, the Science and Technology Development Fund, Macau SAR (Grant no. 0075/2023/AMJ, 0092/2024/AFJ, 0003/2023/RIB1, 001/2024/SKL), the Guangdong Department of Science and Technology (Grant no. 2023A0505030003), and the University of Macau (Grant No.: MYRG-GRG2024-00299-FST-UMDF, MYRG-GRG2025-00312-FST).

\bibliographystyle{IEEEtran}
\bibliography{main}
\clearpage
\appendices
\section{Training Configuration}
\label{app:training}
Table~\ref{tab:hyperparameters} lists settings for driving-domain pretraining, paired SFT, and GRPO. Training uses NVIDIA H20 GPUs, PyTorch 2.5.0, and CUDA 12.3 on Ubuntu; inference uses 8 GPUs.
\begin{table}[H]
\centering\footnotesize\setlength{\tabcolsep}{3pt}\renewcommand{\arraystretch}{1.0}
\caption{Training hyperparameters. A dash indicates a setting not used in supervised training.}
\label{tab:hyperparameters}
\begin{tabular}{@{}lccc@{}}
\toprule
\rowcolor{TableHead} Setting & Stage 1 & Stage 2 & Stage 3 \\
\midrule
Epochs & 2 & 2 & 2 \\
Batch size & 1 & 2 & 4 \\
Learning rate & $1\times10^{-5}$ & $4\times10^{-5}$ & $2\times10^{-6}$ \\
Gradient accumulation steps & 4 & 2 & 16 \\
Weight decay & 0.05 & 0.05 & 0.05 \\
Number of GPUs & 16 & 16 & 64 \\
Generations per query & --- & --- & 12 \\
Iterations & --- & --- & 2 \\
Sampling temperature & --- & --- & 1.0 \\
\bottomrule
\end{tabular}
\end{table}
The Adaptive Think confidence threshold is $T=0.9$. The endpoint reward uses the L1 distance of the final trajectory point, with the piecewise thresholds given in the main text. The format reward totals 1.0: 0.5 for correct reasoning/answer tags and 0.5 for a valid trajectory structure.

\section{Input, Output, and Paired Supervision}
\label{app:interface}
\begin{itemize}\raggedright
\item \textbf{Inputs.} A front-view camera image, a navigation command, velocity and acceleration vectors in the form $[x,y]$, and up to three historical poses $[x,y,\mathrm{heading}]$ sampled at 0.5-second intervals, covering at most 1.5 seconds.
\item \textbf{Prediction horizon.} Eight future poses $[x,y,\mathrm{heading}]$ at 0.5-second intervals, covering 4 seconds. The positive $x$-axis points forward; positive $y$ points left and negative $y$ points right.
\item \textbf{Output structure.} The model emits \texttt{<think>...</think>} followed by \texttt{<answer>...</answer>}. In Non-Think mode, the reasoning tags are empty and the answer still contains the predicted trajectory.
\item \textbf{Reasoning scope.} Think-style supervision considers road boundaries, relevant traffic lights, agents that can affect ego behavior, and the resulting future maneuver. The predicted trajectory follows the planned behavior.
\end{itemize}

\begin{figure*}[t]
\centering
\includegraphics[width=0.92\linewidth]{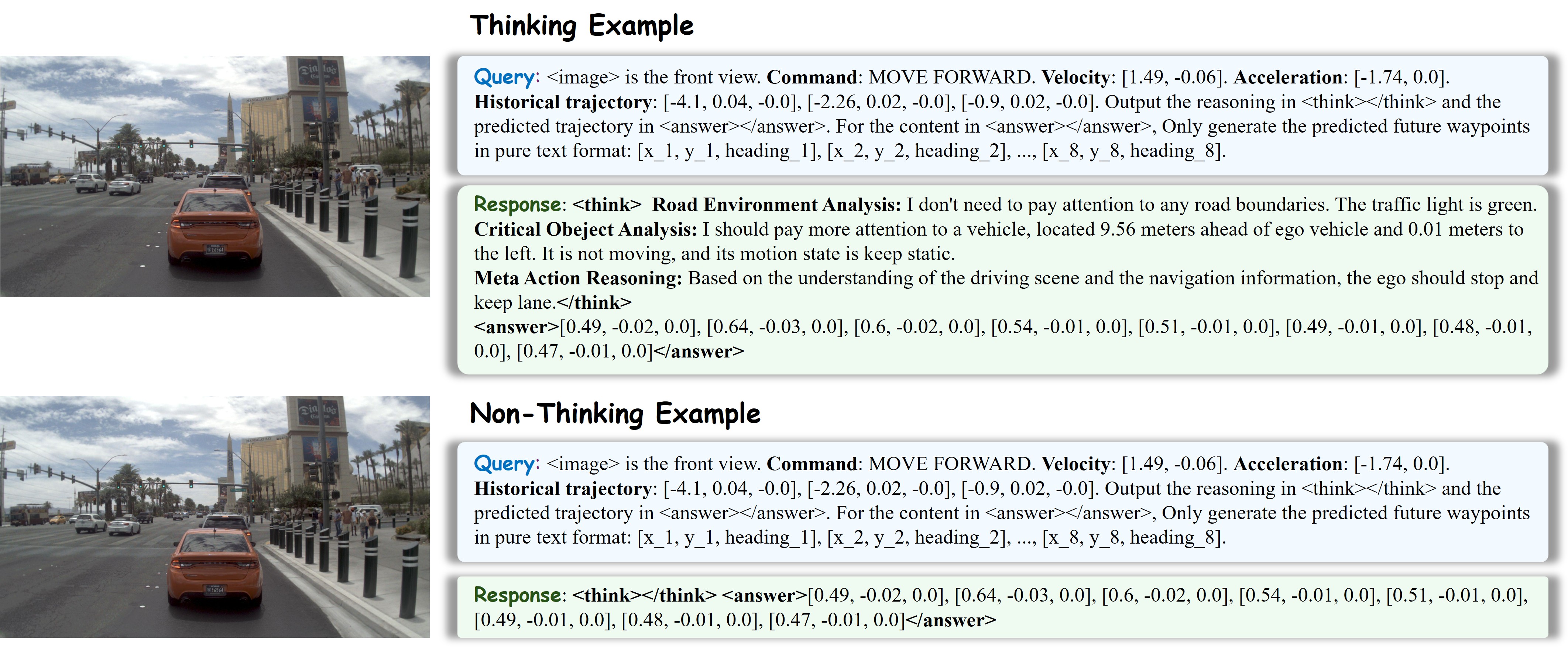}
\caption{Paired SFT supervision: structured reasoning followed by the trajectory (top), and direct trajectory prediction with the same output structure (bottom).}
\label{fig:a2}
\end{figure*}

\paragraph{Paired Supervision}
\label{app:supervision}
Figure~\ref{fig:a2} shows paired outputs for the same visual observation and command. Future trajectories and scene semantics identify relevant agents and boundaries during offline annotation. These signals construct training targets; inference uses only the image, navigation command, ego state, and historical trajectory.

\section{System Prompt}
\label{app:prompt}
The following prompt is used in the SFT and GRPO stages. It specifies the observation fields, the coordinate system, the prediction horizon, and the reasoning/answer structure. Think-style responses include the planning analysis; Non-Think responses leave the reasoning tags empty. The prompt is reproduced in full, with line wrapping normalized for readability.
\begin{tcolorbox}[breakable,title after break={System prompt (continued)},title={System prompt for SFT and GRPO},colback=TableAccent,colframe=Accent,boxrule=0.5pt,arc=2mm,left=4mm,right=4mm,top=3mm,bottom=3mm,fonttitle=\bfseries]
\small
\setlength{\parskip}{6pt}
You are an autonomous driving agent. You have access to the front-view camera image of a vehicle, the drive command, the current velocity and acceleration of the vehicle, and the ego vehicle's historical trajectory.

The drive command represents the desired control command that the system wants the vehicle to maintain. The vehicle's velocity and acceleration are represented as lists [x, y], where x and y correspond to the respective components in the x and y directions. The ego vehicle's historical trajectory consists of past positions in the format [x, y, heading]. It may include up to 3 points, each representing a timestep of 0.5 seconds, with a maximum history of 1.5 seconds.

Your task is to first output the planning reasoning process within the <think> </think> tags, where you must clearly predict the ego vehicle’s future behavior, such as intended maneuvers, path decisions, and interactions with the environment over the next 4 seconds. Then, based strictly on this predicted behavior in the <think> section, generate the corresponding 8-step trajectory in the <answer> </answer> tags. The trajectory must be consistent with and accurately reflect the decisions described in the reasoning process. For the content of answer, you need predict the vehicle’s future trajectory over the next 8 timesteps (covering 4 seconds), based on the surrounding traffic environment observed in the provided images. Each timestep represents 0.5 seconds. Each trajectory point should be a 3D vector: [x, y, heading], where x and y are the 2D coordinates and heading is the orientation angle. The positive y-axis points to the left side of the vehicle, and the negative y-axis points to the right. The positive x-axis points in the forward driving direction. If you fail to provide a response, it may lead to potential harm or injury. You can provide a detailed reasoning process in the <think></think> tag, or the <think></think> tag must remain empty.

The planning reasoning in the <think></think> is as follows: First, examine whether ego vehicle is close to any road boundaries. If it is, explain how the ego should slightly deviate away from the edge to maintain safe driving. If there are no road boundaries to be concerned about, simply respond with: "I don't need to pay attention to any road boundaries." Next, check if traffic lights are visible and relevant. If a traffic light is present, provide guidance on how the ego vehicle should respond (e.g., stop at a red light).  If there is no traffic lights, simply respond with: "There is no traffic light." Then, describe only the agents that may affect the ego vehicle's driving. Focus on their relative distance and motion state. Based on this analysis, reason about the ego vehicle’s next driving behavior. If there are no agents influencing the car's driving behavior, simply respond with: There are no objects in the current scene that require my attention. Regardless of the situation, the model must predict the ego vehicle's future driving behavior.

\end{tcolorbox}

\subsection{Reward Implementation Details}
\label{app:reward-details}
The format reward contributes up to 1.0 point: 0.5 for the required reasoning and answer segments and 0.5 for a valid trajectory representation. The endpoint reward uses the L1 distance $d$ between the final predicted and reference positions (in meters):
\begin{equation}
\mathcal R_{\mathrm{endpoint}}(d)=
\begin{cases}
1.0,&0\leq d<2,\\
0.8,&2\leq d<4,\\
0.6,&4\leq d<6,\\
0.4,&6\leq d<10,\\
0.2,&10\leq d<15,\\
0.0,&d\geq15.
\end{cases}
\end{equation}
These settings follow the main-paper configuration: RL batch size 4, continuous PDMS reward in $[0,1]$, and Adaptive Think threshold $T=0.9$.

\section{Additional Qualitative Examples}
\label{app:qualitative}
\subsection{Driving-Domain Question Answering}
\begin{figure*}[t]
\centering
\begin{subfigure}{\linewidth}
\includegraphics[width=\linewidth]{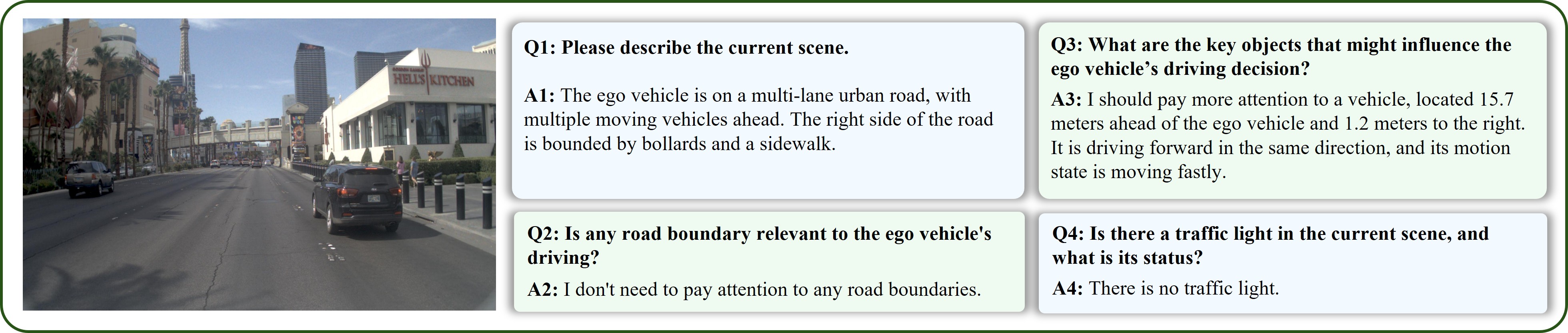}
\caption{Driving-domain question answering: example 1.}\label{fig:vis1}
\end{subfigure}\par\vspace{8pt}
\begin{subfigure}{\linewidth}
\includegraphics[width=\linewidth]{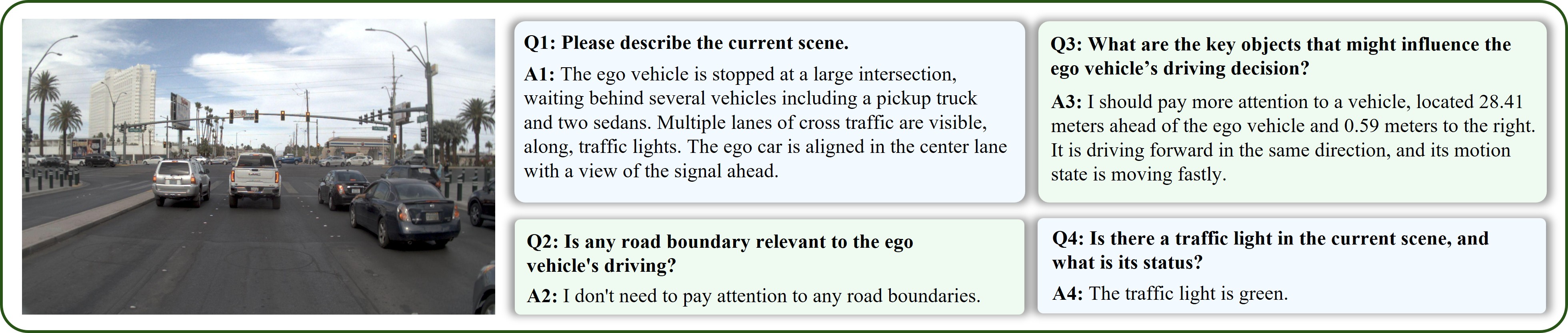}
\caption{Driving-domain question answering: example 2.}\label{fig:vis2}
\end{subfigure}\par\vspace{8pt}
\begin{subfigure}{\linewidth}
\includegraphics[width=\linewidth]{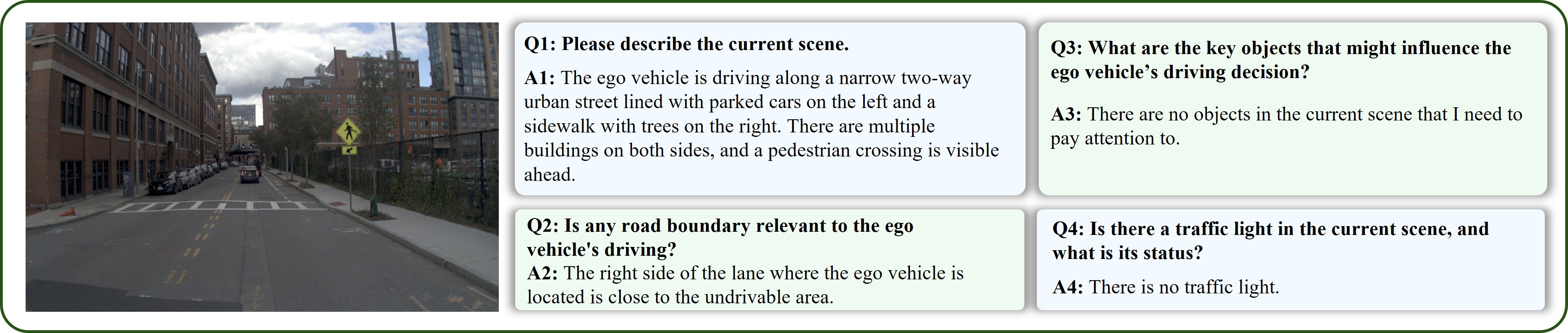}
\caption{Driving-domain question answering: example 3.}\label{fig:vis3}
\end{subfigure}
\caption{Qualitative reasoning examples from the pretraining stage, covering scene understanding, road-boundary relevance, traffic lights, and critical-object identification.}
\label{fig:appendix-domain-qa}
\end{figure*}

Figure~\ref{fig:appendix-domain-qa} shows three pretraining-stage examples. Each example contains an observation and answers about scene description, relevant road boundaries, traffic light status, and key traffic participants. These qualitative examples supplement the paired planning supervision illustrated in Figure~\ref{fig:a2}.

\subsection{Planning in Simple and Challenging Scenes}
\begin{figure*}[t]
\centering
\begin{subfigure}{\linewidth}
\includegraphics[width=\linewidth]{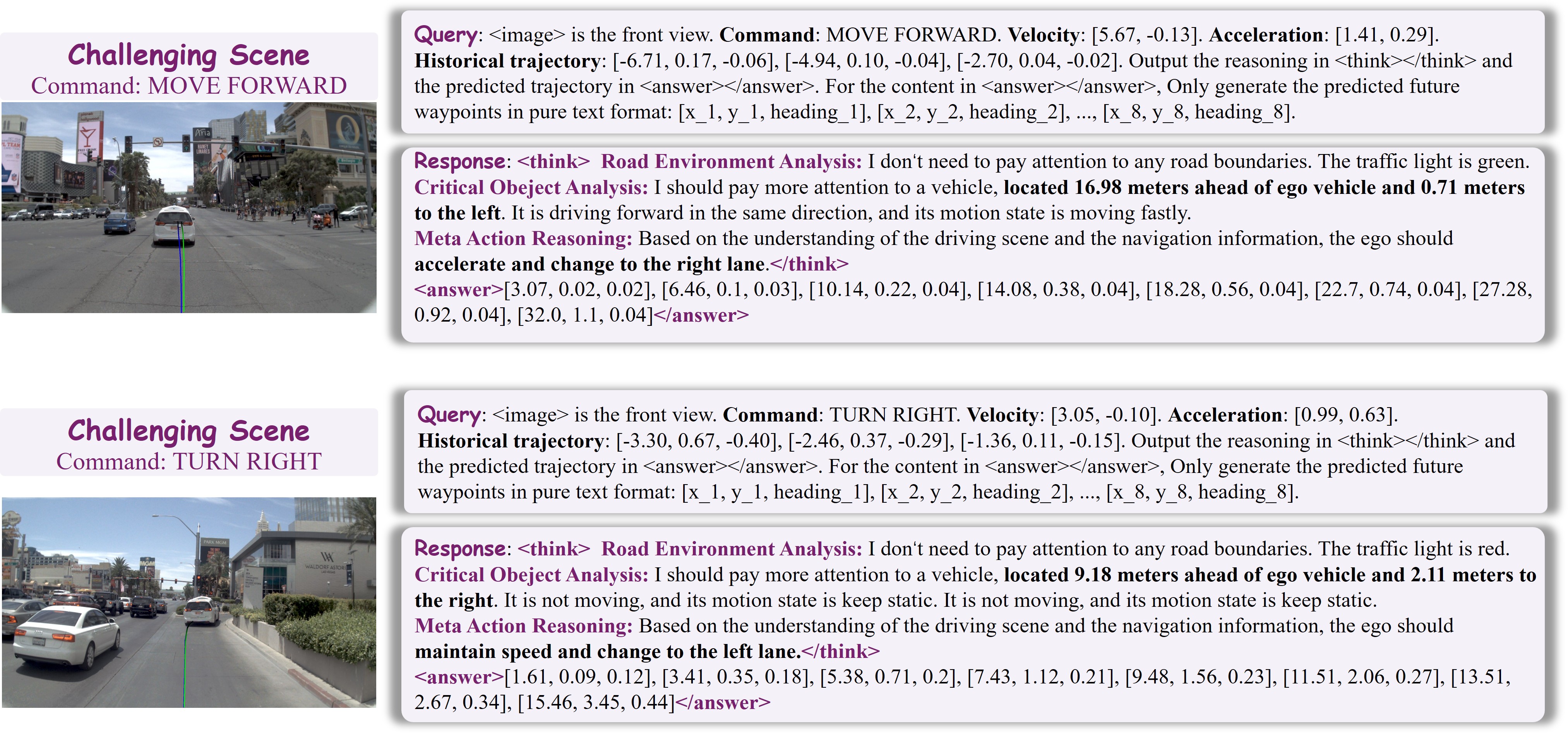}
\caption{Challenging scenes: reasoning followed by trajectory prediction.}\label{fig:a3}
\end{subfigure}\par\vspace{10pt}
\begin{subfigure}{\linewidth}
\includegraphics[width=\linewidth]{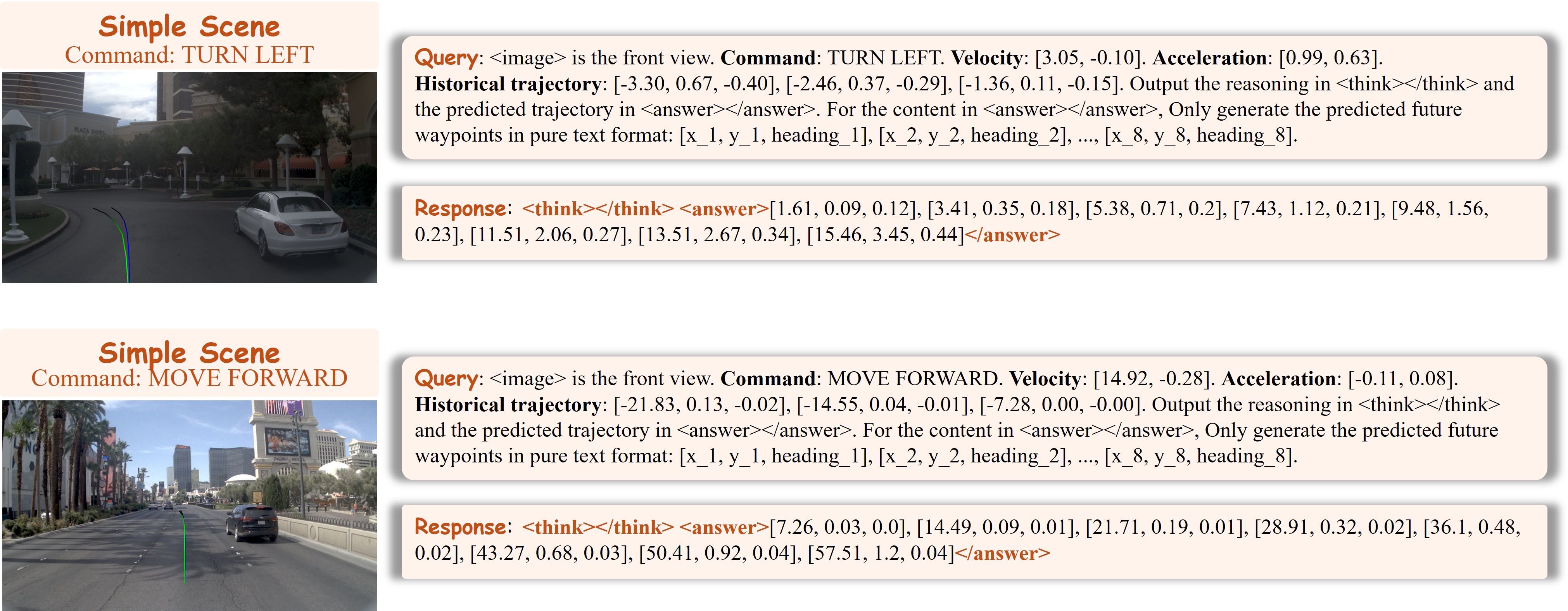}
\caption{Simple scenes: direct trajectory prediction.}\label{fig:a4}
\end{subfigure}
\caption{Additional adaptive-planning examples after reinforcement learning. Each case includes the observation, navigation command, ego state and history, model response, and predicted trajectory.}
\label{fig:appendix-planning}
\end{figure*}

Figure~\ref{fig:appendix-planning} provides additional RL-stage examples. Challenging scenes include structured road-environment, critical-object, and maneuver analysis before the trajectory. The simple-scene examples directly produce trajectories with an empty reasoning segment. Green and blue denote the reference and predicted trajectories, respectively.

\end{document}